\def\year{2016}\relax
\pdfoutput=1
\documentclass[letterpaper]{article}
\usepackage{aaai16}
\usepackage{times}
\usepackage{helvet}
\usepackage{courier}
\usepackage{graphicx}
\usepackage{subfigure}
\begin{document}
% The file aaai.sty is the style file for AAAI Press
% proceedings, working notes, and technical reports.
%
\title{Building a Large Scale Dataset for Image Emotion Recognition: The Fine Print and The Benchmark}
\author{Quanzeng You \and Jiebo Luo \\
Department of Computer Science \\
University of Rochester \\
Rochester, NY 14623 \\
\{qyou, jluo\}@cs.rochester.edu \\
\And Hailin Jin \\
Adobe Research \\
345 Park Avenue\\
San Jose, CA 95110\\
hljin@adobe.com \\
\And Jianchao Yang \\
Snapchat Inc\\
64 Market St\\
Venice, CA 90291\\
jianchao.yang@snapchat.com\\
}
\maketitle
\begin{abstract}
\begin{quote}
Psychological research results have confirmed that people can have different emotional reactions to different visual stimuli. Several papers have been published on the problem of visual emotion analysis. In particular, attempts have been made to analyze and predict people's emotional reaction towards images. To this end, different kinds of hand-tuned features are proposed. The results reported on several carefully selected and labeled small image data sets have confirmed the promise of such features. While the recent successes of many computer vision related tasks are due to the adoption of Convolutional Neural Networks (CNNs), visual emotion analysis has not achieved the same level of success. This may be primarily due to the unavailability of confidently labeled and relatively large image data sets for visual emotion analysis. In this work, we introduce a new data set, which started from 3+ million weakly labeled images of different emotions and ended up 30 times as large as the current largest publicly available visual emotion data set. We hope that this data set encourages further research on visual emotion analysis. We also perform extensive benchmarking analyses on this large data set using the state of the art methods including CNNs.
\end{quote}
\end{abstract}

\section{Introduction}
Psychological studies have provided evidence that human emotions can be aroused by visual content, e.g. images~\cite{lang1979bio,lang1998emotion,joshi2011aesthetics}. Based on these findings, recently computer scientists also started to delve into this research topic. However, differently from psychological studies, which mainly focus on studying the changes between physiological and psychological activities of human beings on visual stimuli, most of the works in computer science are trying to predict the aroused human emotion given a particular piece of visual content. Indeed, affective computing, which \textit{aims to recognize, interpret and process human affects} (http://en.wikipedia.org/wiki/Affective\_computing), has achieved significant progress in recent years. However, the problem of visual emotion prediction is more difficult in that we are trying to predict the emotional reactions given a general visual stimulus, instead of using the collected signals from human's physiological reactions of visual stimuli as studied in affective computing.
\begin{figure}
\begin{center}
\includegraphics[width=.45\textwidth]{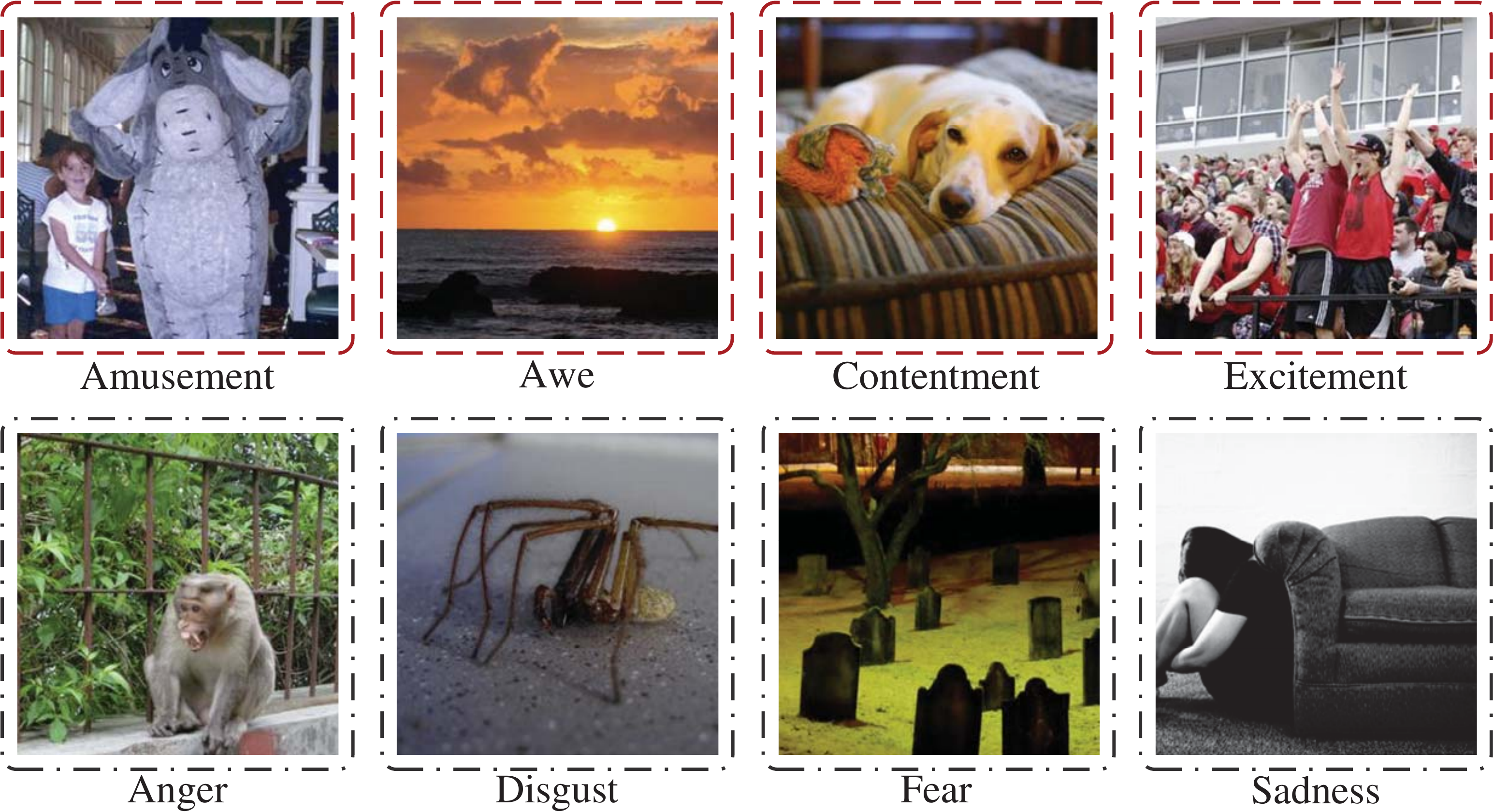}
\end{center}
\caption{Example images of eight different emotion categories. \textit{Top Row}: four positive emotions and \textit{Bottom Row}: four negative emotions.}
\label{fig:example:imgs}
\end{figure}

The increasing popularity of social networks attracts more and more people to publish multimedia content in online social network platforms.
%Each day, billions of messages and posts are generated.
Online users can easily add textual data, e.g., title, descriptions, tags, to their uploaded images and videos. However, these textual information can only help the retrieval of multimedia content in the cognitive level~\cite{machajdik2010affective}, i.e., semantic level. The meta text data has limited help in bridging the \textit{affective} semantic gap between images pixels and human feelings. In~\cite{hanjalic2006extracting}, the authors call visual emotion prediction \textit{affective content analysis}.

Inspired by the psychology and art theory, different groups of manually crafted features are designed to study the emotional reactions towards visual content. For example, based on art theory~\cite{itten1973art,valdez1994effects}, Machajdik and Hanbury~\cite{machajdik2010affective} defined eight different kinds of pixel level features (e.g. color, texture and composition), which have been empirically proved to be related to emotional reactions.  In another recent work~\cite{Zhao:2014:EPF:2647868.2654930}, \textit{principles-or-art} based features are extracted to classify emotions. Following their works, we study the same eight emotions, \textit{Amusement}, \textit{Awe}, \textit{Contentment}, \textit{Excitement}, \textit{Anger}, \textit{Disgust}, \textit{Fear} and \textit{Sadness}. \figurename~\ref{fig:example:imgs} shows the example images for these studied emotions. All these images are selected from the newly constructed data set in this work, where each image is labeled by five Amazon Mechanical Turk workers.

Recently deep learning has enabled robust and accurate feature learning, which in turn produces the state-of-the-art performance on many computer vision related tasks, e.g.~digit recognition~\cite{lecun1989backpropagation,hinton2006fast}, image classification~\cite{cirecsan2011flexible,krizhevsky2012imagenet}, aesthetics estimation~\cite{lu2014rapid} and scene recognition~\cite{zhou2014learning}. One of the main factors that prompt the success of deep learning on these problems is the availability of a large scale data set. From ImageNet~\cite{deng2009imagenet} to AVA dataset~\cite{murray2012ava} and the very recent Places Database~\cite{zhou2014learning}, the availability of these data sets have significantly promoted the development of new algorithms on these research areas. %In particular, to train a deep neural network, one needs to have a well labeled data set.
In visual emotion analysis, there is no such a large data set with strong labels. More recently, You~et al.~\cite{you2015robust} employed CNNs to address visual sentiment analysis, which tries to bridge the high-level, abstract sentiments concept and the image pixels. They employed the \textit{weakly} label images to train their CNN model. However, they are trying to solve a binary classification problem instead of a multi-class (8 emotions) problem as studied in this work. %Their experimental results suggest that CNNs outperform other manually designed low-level and mid-level (concept and attribute level) features on visual sentiment analysis.

In this work, we are interested in investigating the possibility of solving the challenging visual emotion analysis problem. First, we build a large scale emotion data set. On top of the data set, we intend to find out whether or not applying CNNs to visual emotion analysis provides advantages over using a predefined collection of art and psychology theory inspired visual features or visual attributes, which have been done in prior works. To that end, we make the following contributions in this work.

\begin{itemize}
\item{We collect a large number of weakly labeled emotion related images. Next, we employ Amazon Mechanical Turk to manually label these images to obtain a relatively strongly labeled image data set, which makes the usage of CNN for visual emotion analysis possible. All the data set will be released to the research community upon publishing this work.}
\item{We evaluate the performance of Convolutional Neural Networks on visual emotion analysis and establish it as the baseline for future research. Compared with the state-of-the-art manually crafted visual features, our results suggest that using CNN can achieve significant performance improvement on visual emotion analysis.}
\end{itemize}
\section{Related Work}
\label{sec:related}
Our work is mostly related to both visual emotion analysis and Convolutional Neural Networks (CNNs). Recently, deep learning has achieved massive success on a wide range of artificial intelligence tasks. In particular, deep Convolutional Neural Networks have been widely employed to solve traditional computer vision related problems. Deep convolutional neural networks typically consist of several convolutional layers and several fully connected layers. Between the convolutional layers, there may also be pooling layers and normalization layers. In early studies, CNNs~\cite{lecun1998gradient} have been very successful in document recognition, where the inputs are relatively small images. Thanks to the increasing computational power of GPU, it is now possible to train a deep convolutional neural network on large collections of images (e.g.~\cite{krizhevsky2012imagenet}), to solve other computer vision problems, such as scene parsing~\cite{grangier2009deep}, feature learning~\cite{lecun2010convolutional}, visual recognition~\cite{kavukcuoglu2010learning} and image classification~\cite{krizhevsky2012imagenet}.

%Besides those traditional vision related tasks, CNNs are also applied to other high-level visual content related tasks. Karayev~et al.~\cite{karayev2013recognizing} apply CNN for the recognition of image styles. They find that deep visual features are generally better than hand-tuned features in recognizing image styles. Besides image style, aesthetics of images are estimated in Lu~et al.~\cite{lu2014rapid} by applying CNN to automatic feature learning. More recently, deep visual features are employed in~\cite{you2015robust} for visual sentiment analysis. Their results show that even using the \textit{weakly} labeled images along with a progressive training strategy, the trained CNN models can outperform the state-of-the-art low-level and mid-level visual features for visual sentiment analysis. Therefore in this study, we focus on using CNN to establish baselines for this new image emotion dataset.

\begin{table*}
\small
\begin{center}
\begin{tabular}{*{10}{|c}|}
\hline
Data Set & Amusement & Anger & Awe & Contentment & Disgust & Excitement & Fear & Sadness & Sum \\
\hline\hline
IAPS-Subset & 37& 8& 54& 63 & 74& 55&42&62 & 395 \\
ArtPhoto & 101 &77&102&70&70& 105 & 115 & 166 & 806 \\
Abstract Paintings & 25 & 3 & 15 & 63&  18 & 36 & 36 & 32 & 228 \\
In Total & 163 & 88 & 171 & 196 & 162 & 196 & 193 & 260 & 1429 \\
\hline
\end{tabular}
\end{center}
\caption{Statistics of the three existing data sets. The three data sets are imbalanced across the 8 categories.}
\label{tab:dataset:base}
\end{table*}
However, to the best of our knowledge, there are no related works on using CNNs for visual emotion analysis. Currently, most of the works on visual emotion analysis can be classified into either dimensional approach~\cite{nicolaou2011multi,Lu:2012:SCE:2393347.2393384} or categorical approach~\cite{machajdik2010affective,borth2013large,Zhao:2014:EPF:2647868.2654930}, where the former represents emotion in a continuous two dimensional space and in the later model each emotion is a distinct class. We focus on the categorical approach, which has been studied in several previous works. Jia~et al.~\shortcite{jia2012can} extract \textit{color} features from the images. With additional social relationships, they build a factor graph model for the prediction of emotions. Inspired by art and psychology theory, Machajdik and Hanbury~\shortcite{machajdik2010affective} proposed richer hand-tuned features, including \textit{color}, \textit{texture}, \textit{composition} and \textit{content} features. Furthermore, by exploring the principles of art, Zhao~et al.~\shortcite{Zhao:2014:EPF:2647868.2654930} defined more robust and invariant visual features, such as \textit{balance}, \textit{variety}, and \textit{gradation}. Their features achieved the best reported performance on several publicly accessible emotion data sets.

Those hand-tuned visual features have been validated on several publicly available small data sets (see following sections for details). However, we want to verify whether or not deep learning could be applied to this challenging problem and more importantly, on a much larger scale image set. The main issue is that there are no available well labeled data sets for training deep neural networks. Our work intends to provide such a data set for the research community and verify the performance of the widely used deep convolutional neural architecture on this emotion data set.
\section{Visual Emotion Data Sets}
\label{sec:dataset}

Several small data sets have been have been used for visual emotion analysis~\cite{yanulevskaya2008emotional,machajdik2010affective,Zhao:2014:EPF:2647868.2654930}, including (1) \textbf{IAPS-Subset:} This data set is a subset of the International Affective Picture System (IAPS)~\cite{lang1999international}. This data set is categorized into eight emotional categories as shown in~\figurename~\ref{fig:example:imgs} in a study conducted in~\cite{mikels2005emotional}. (2) \textbf{ArtPhoto:} Machajdik and Hanbury~\shortcite{machajdik2010affective} built this data set, which contains photos by professional artists. They obtain the ground truth by the labels provided by the owner of each image. (3) \textbf{Abstract Paintings:} These are images consisting of both color and texture from~\cite{machajdik2010affective}. They obtain the ground truth of each image by asking people to vote for the emotions of each image in the given eight emotion categories. \tablename~\ref{tab:dataset:base} shows the statistics of the existing three data sets. The numbers show that each data set only consists of a very small number of images. Meanwhile, images in all the three different data sets are highly imbalanced. All three data sets are relatively small with images coming from a few specific domains. In particular, for some categories, such as \textit{Anger}, there are less than $10$ images. Therefore, if we employ the same methodology (5-fold Cross Validation within each data set)~\cite{machajdik2010affective,Zhao:2014:EPF:2647868.2654930}, we may have only several images in the training data. This may lead to the possibility that their trained models may have been either over or under fitted.
%\begin{figure}[htbp]
%\begin{center}
%\includegraphics[width=.45\textwidth]{./figures/emotion_baseline}
%\end{center}
%\caption{Embedding of the three data sets using features extracted using CNN model trained on ImageNet. We use t-SNE~\cite{van2008visualizing} to reduce the features into a 2-d dimension.}
%\label{fig:embedding:base}
%\end{figure}

%\figurename~\ref{fig:embedding:base} shows the images from the three data sets. We employ the pre-trained CNN model on ImageNet~\cite{krizhevsky2012imagenet,jia2014caffe} to extract the visual features for images in the three data sets using the second to the last layer. Next, we employ t-SNE~\cite{van2008visualizing}, which is widely used for visualization of high-dimensional data sets, to embed all the images in the three data sets into a 2-d space. From this embedding, we can see that images are kind of separated by the content they carry. For instance, most images in the right part come from images in the \textit{Abstract Paintings} and most images in the left up corner are related to natural scenes. This may be partially due to the construction of semantic features from object-oriented deep ImageNet models.

The above results suggest that the previous efforts on visual emotion analysis deal with small emotion-centric data sets compared with other vision data sets, such as ImageNet~\cite{deng2009imagenet} and Places~\cite{zhou2014learning}. In this work, we present here a new emotion data set, which is by far the largest available emotion-centric database.
\subsection{Building An Image Emotion Dataset from the Wild}
\begin{figure}
\begin{center}
\includegraphics[width=.4\textwidth]{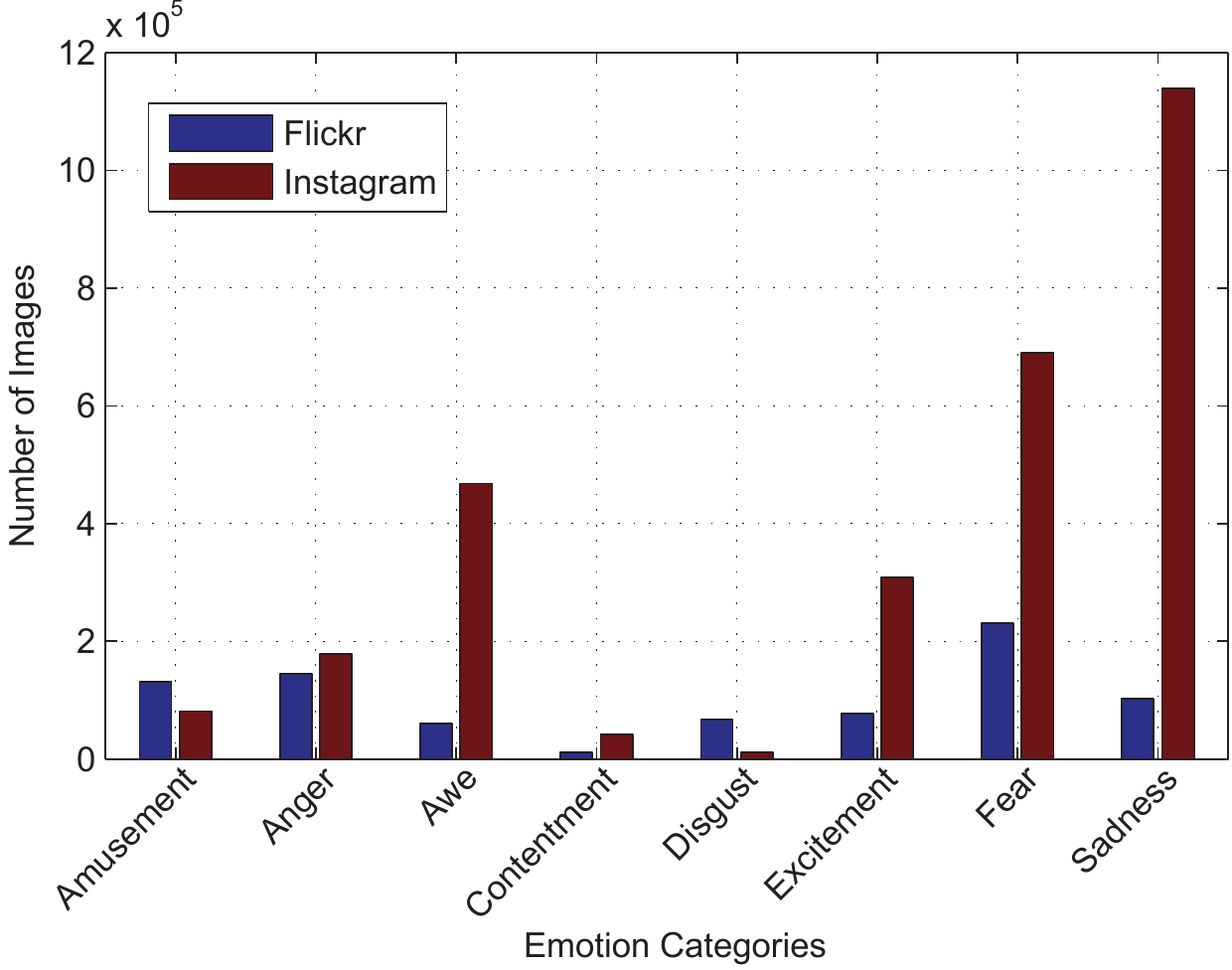}
\end{center}
\caption{Statistics of the images downloaded from Flickr and Instagram.}
\label{fig:stat}
\end{figure}

There are many different emotion definition systems\footnote{http://en.wikipedia.org/wiki/Emotion} from psychological and cognitive science. In this work, we use the same eight emotions defined in~\tablename~\ref{tab:dataset:base}, which is derived from a psychological study in~\cite{mikels2005emotional}.
Using the similar approach in~\cite{jia2012can}, we query the image search engines (Flickr and Instagram) using the eight emotions as keywords. In this way, we are able to collect a total of over 3 million \textit{weakly} labeled images, i.e., labeled by the queries. Next, we delete images which have tags of any two different emotions. We also remove duplicate images using \textit{fdupes}\footnote{https://code.google.com/p/fdupes/}. \figurename~\ref{fig:stat} shows the statistics of the remaining images. As we can see, the number of images in different categories is imbalanced. In particular, there are only small numbers of \textit{contentment} and \textit{disgust} images in both social media platforms. Meanwhile, the number of per category images from Instagram varies significantly. There are much more images from both \textit{Fear} and \textit{Sadness}. This agrees with the finding (http://goo.gl/vhBBF6) that people are likely to share sadness from their Instagram accounts.

Next, we employ Amazon Mechanical Turk (AMT) to further label these \textit{weakly} labeled images. In particular, we design a qualification test to filter all workers who want to work on our tasks. The qualification test is designed as an image annotation problem. We randomly select images from the publicly available \textit{ArtPhoto} data set and use the ground-truth labels as the answers. For each given image, we ask the workers to choose the emotion they feel from the eight emotion categories. At first, we conduct experiments within members in our research group. Indeed, the results suggest that this qualification is challenging difficult, in particular when you have to choose only one emotion for each image. Therefore, we design our AMT tasks (HITs) as a much easier verification task instead of the annotation task. Since we have crawled all the images with emotion queries, we have a \textit{weakly} labeled data set. In each HIT, we assign \textit{five} AMT workers to verify the emotion of each image. For each given image and its \textit{weak} label, we ask them to answer a question like \textit{ Do you feel anger seeing this image?} The workers are asked to choose a YES or NO for each image. All workers have to meet the rigorous requirement of correctly answering at least half of the questions (a total of 20) in our qualification test. By the time of finishing work, we have over $1000$ workers on our qualification task. Among them, $225$ workers meet our qualification criteria.

To start the verification task, we randomly select $11,000$ images for each emotion category. After collecting the batch results from AMT, we keep those images which receive at least three Yeses from their assigned fiver AMT workers. In this way, we are able to build a relatively strongly labeled data set for visual emotion analysis.
\tablename~\ref{tab:dataset:our} summarizes the number of images for our current data set. The numbers show that different categories may have different acceptance rates for workers to reject or accept the \textit{positive} samples of each verification task. In particular, we add another $2,000$ images to make sure that the number of images in \textit{Fear} category is also larger than $1,000$ images. In total, we collect about $23,000$ images, which is about $30$ times as large as the current largest emotion data set, i.e., \textit{ArtPhoto}.
\begin{table*}[!htpb]
\begin{center}
\begin{tabular}{*{10}{|c}|}
\hline
Data Set & Amusement & Anger & Awe & Contentment & Disgust & Excitement & Fear & Sadness & Sum \\
\hline\hline
Submitted & 11,000& 11,000& 11,000& 11,000& 11,000& 11,000& 13,000& 11,000 & 90,000 \\ \hline
Labeled & 4,942&1,266&3,151&5,374&1,658&2,963&1,032&2,922 & 23,308 \\
\hline
\end{tabular}
\end{center}
\caption{Statistics of the current labeled image data set. Note that all emotion categories have 1000+ images.}
\label{tab:dataset:our}
\end{table*}

\section{Fine-tuning Convolutional Neural Network for Visual Emotion Analysis}
\label{sec:cnn}
%\begin{figure}[!t]
%\begin{center}
%\subfigure[Filters of CONV1 Layer from ImageNet-CNN]{
%\includegraphics[width=.45\textwidth]{./conv1_weight-crop3}
%}
%\subfigure[Filters of CONV1 Layer from Noisy-Fine-tuned-CNN]{
%\includegraphics[width=.45\textwidth]{./conv1_flickr_ins_weight-crop3}
%}
%\subfigure[Filters of CONV1 Layer from Fine-tuned-CNN]{
%\includegraphics[width=.45\textwidth]{./conv1_weight_emotion-crop3}
%}
%\end{center}
%\caption{Visualization of learned filters by the three different deep models.}
%\label{fig:cnn:filters}
%\end{figure}
Convolutional Neural Networks (CNN) have been proven to be effective in image classification tasks, e.g., achieving the state-of-the-art performance in ImageNet Challenge~\cite{krizhevsky2012imagenet}. Meanwhile, there are also successful applications by fine-tuning the pre-trained ImageNet model, including recognizing image style~\cite{karayev2013recognizing} and semantic segmentation~\cite{long2014fully}. In this work, we employ the same strategy to fine-tune the pre-trained ImageNet reference network~\cite{jia2014caffe}. The same neural network architecture is employed. We only change the last layer of the neural network from $1000$ to $8$. The remain layers keep the same as the ImageNet reference network. We randomly split the collected $23,000$ samples into training (80\%), testing (15\%) and validating sets (5\%).

Meanwhile, we also employ the \textit{weak} labels~\cite{jia2012can} to fine-tune another model as described in~\cite{you2015robust}. We exclude those images that have been chosen to be submitted to AMT for labeling. Next, since \textit{contentment} contains only about $16,000$ images, we randomly select $20,000$ images for other emotion categories. In this way, we have a total of $156,000$ images. We call this model \textit{Noisy-Fine-tuned CNN}. We fine-tune both models using Caffe with a Linux server with 2 NVIDIA TITAN GPUs on top of the pre-trained ImageNet CNN model. %\figurename~\ref{fig:cnn:filters} shows the learned filters of the 1st-convolutional layer. To make a relative comparison, we normalize the filter weights of the three models together before the visualization. The figure shows that the structures in most of the filters seem to be quite similar. However, the contrasts are quite different from filters to filters, which may be attributed to fine-tuning the model using images from different domains. %In particular, there are more colorful filters from the fine-tuned CNN, which is related to the finding that emotions are correlated with colors~\cite{hemphill1996note}.

\subsection{Performance of Convolutional Neural Networks on Visual Emotion Analysis}
After the fine-tuning of the pre-trained CNN model, we obtain two new CNN models. To compare with the ImageNet-CNN, we also show the results of using the SVM trained on features extracted from the second to the last layer of the pre-trained ImageNet-CNN model. In particular, we employ PCA to reduce the dimensionality of the features. We also try several different numbers of principal components. The results are almost the same. To overcome the imbalance problem in the data, we adjust the weights of SVM for different classes (in our implementation, we use LIBSVM\footnote{http://www.csie.ntu.edu.tw/$\sim$cjlin/libsvm/}, which provides such a mechanism). \tablename~\ref{tab:testing:acc} summarizes the performance of the three groups of features on the $15\%$ randomly chosen testing data. The overall accuracy of the Fine-tuned-CNN is almost 60\%. As a baseline, the visual features extracted from ImageNet-CNN only lead to an overall accuracy of about $30\%$, which is half of Fine-tuned-CNN. The Noisy-Fine-tuned-CNN model has an overall accuracy of about $46\%$, which suggests that this model can learn some knowledge from the noisily labeled images. However, even though it has much more training samples compared with \textit{Fine-tuned CNN}, it fails to outperform \textit{Fine-tuned CNN}, which is trained on strongly labeled samples.

\begin{figure*}[!t]
\begin{center}
\subfigure[ImageNet-CNN (avg: 0.28)]{
\includegraphics[width=.3\textwidth]{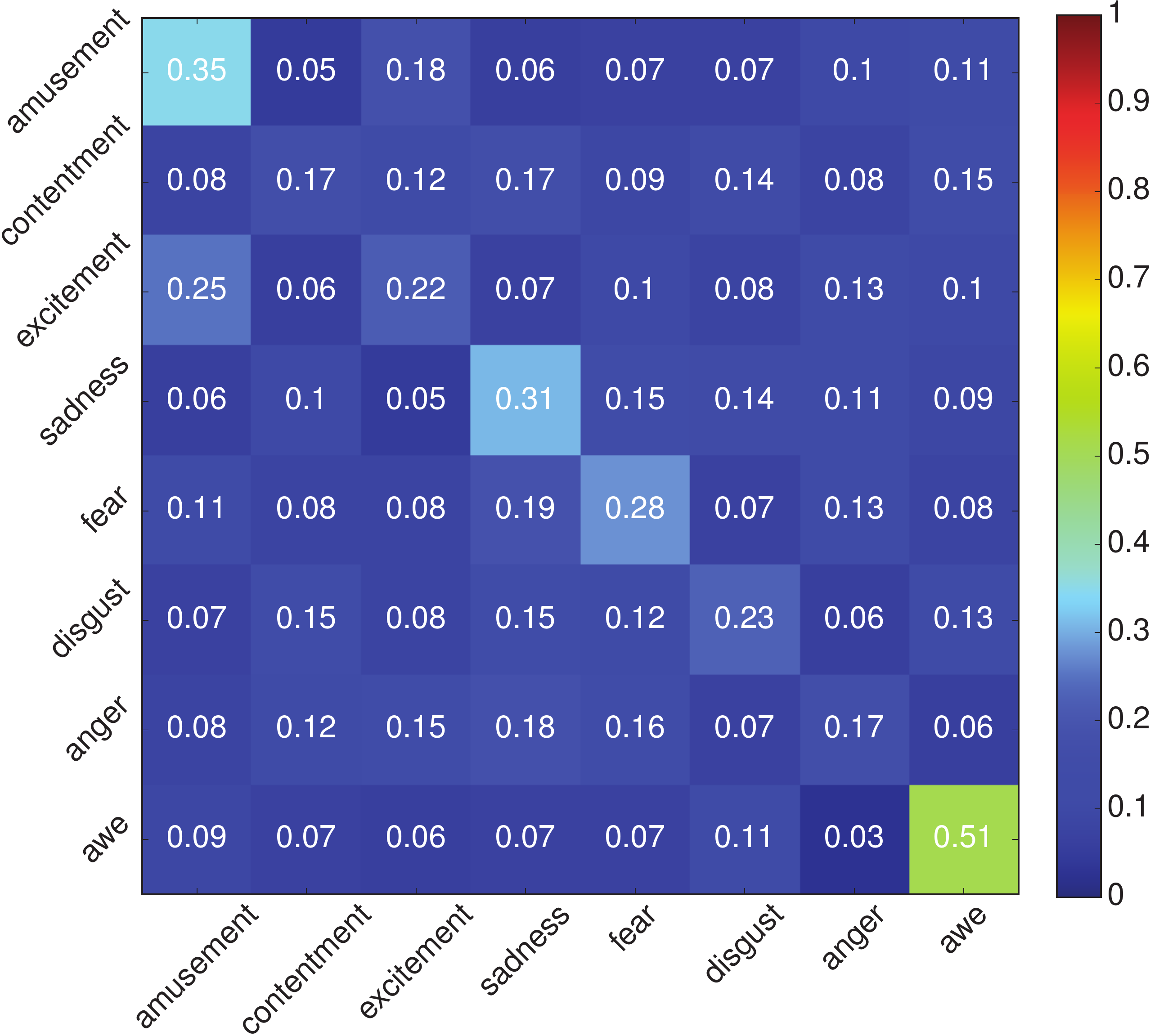}
\label{fig:confusion:imagenet}
}
\subfigure[Noisy-Fine-tuned-CNN (avg: 0.459)]{
\includegraphics[width=.3\textwidth]{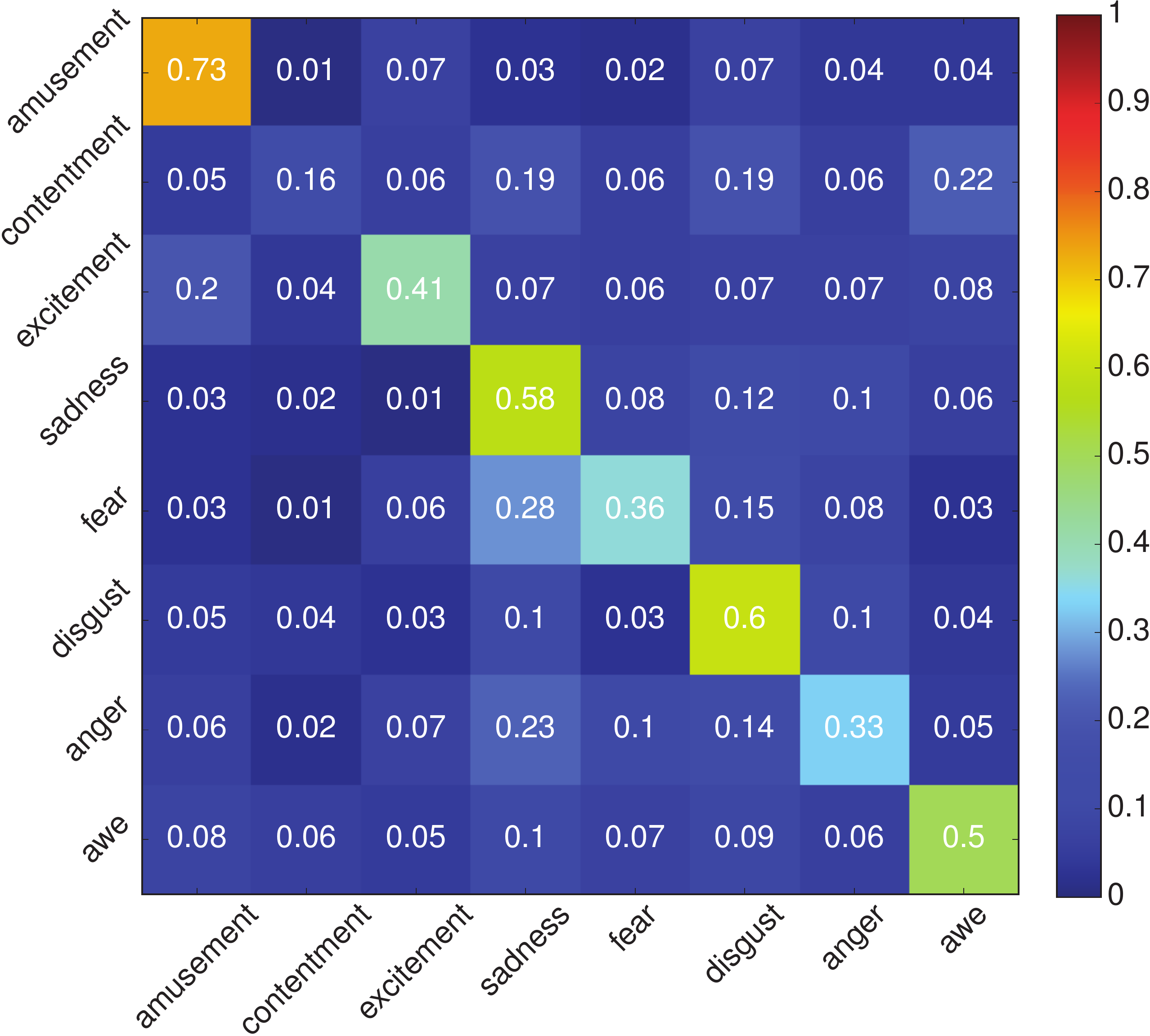}
\label{fig:confusion:noisyfinetune}
}
\subfigure[Fine-tuned-CNN (avg: 0.483)]{
\includegraphics[width=.3\textwidth]{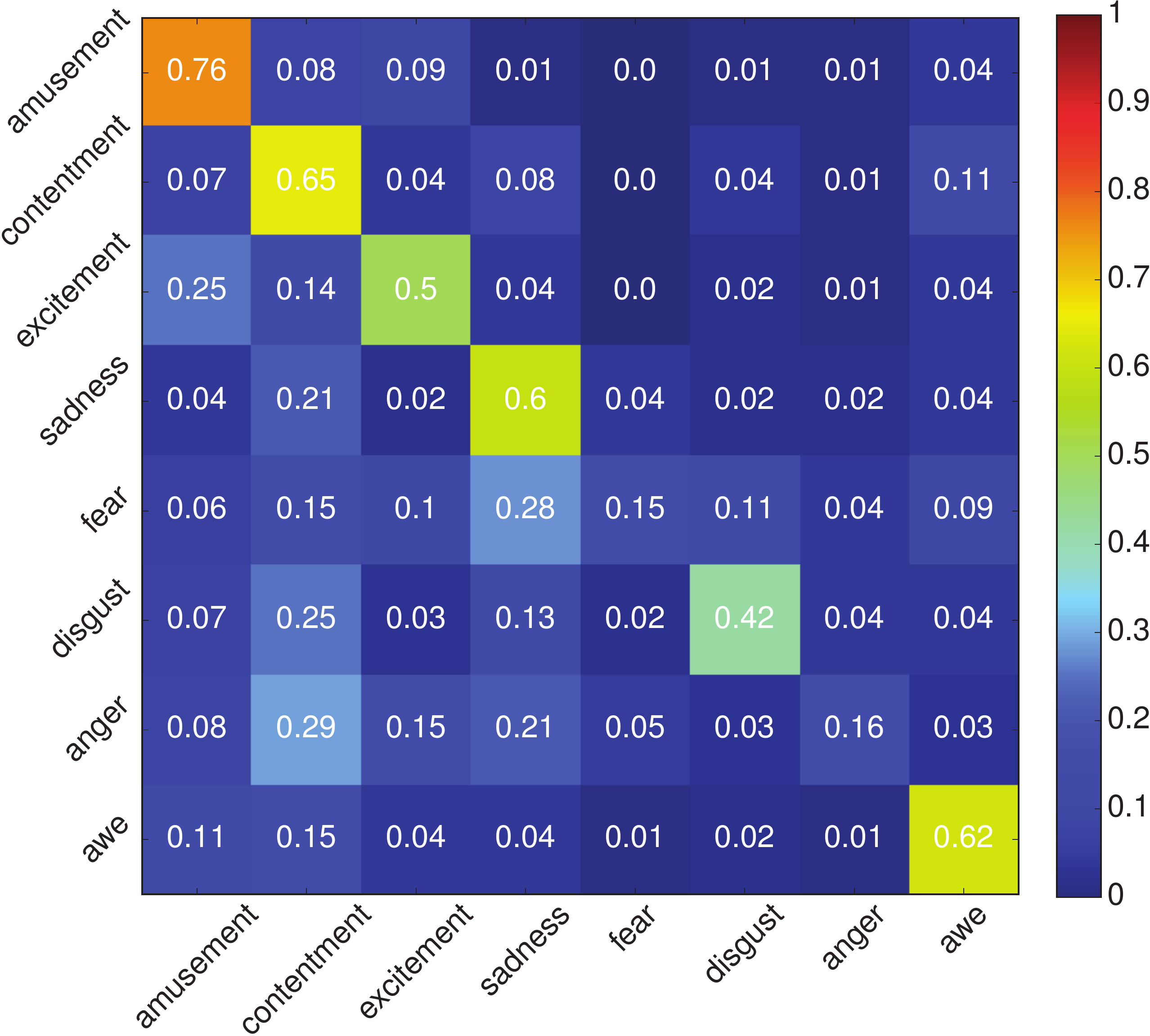}
\label{fig:confusion:finetune}
}
\end{center}
\caption{Confusion matrix for ImageNet-CNN, Noisy-Fine-tuned-CNN and Fine-tuned-CNN on the testing Amazon Mechanical Turk (AMT) labeled images.}
\label{fig:confusion}
\end{figure*}

\begin{figure*}[!t]
\begin{center}
\subfigure[Embedding of the \textit{testing} images using general ImageNet-CNN features]{
\includegraphics[width=.4\textwidth]{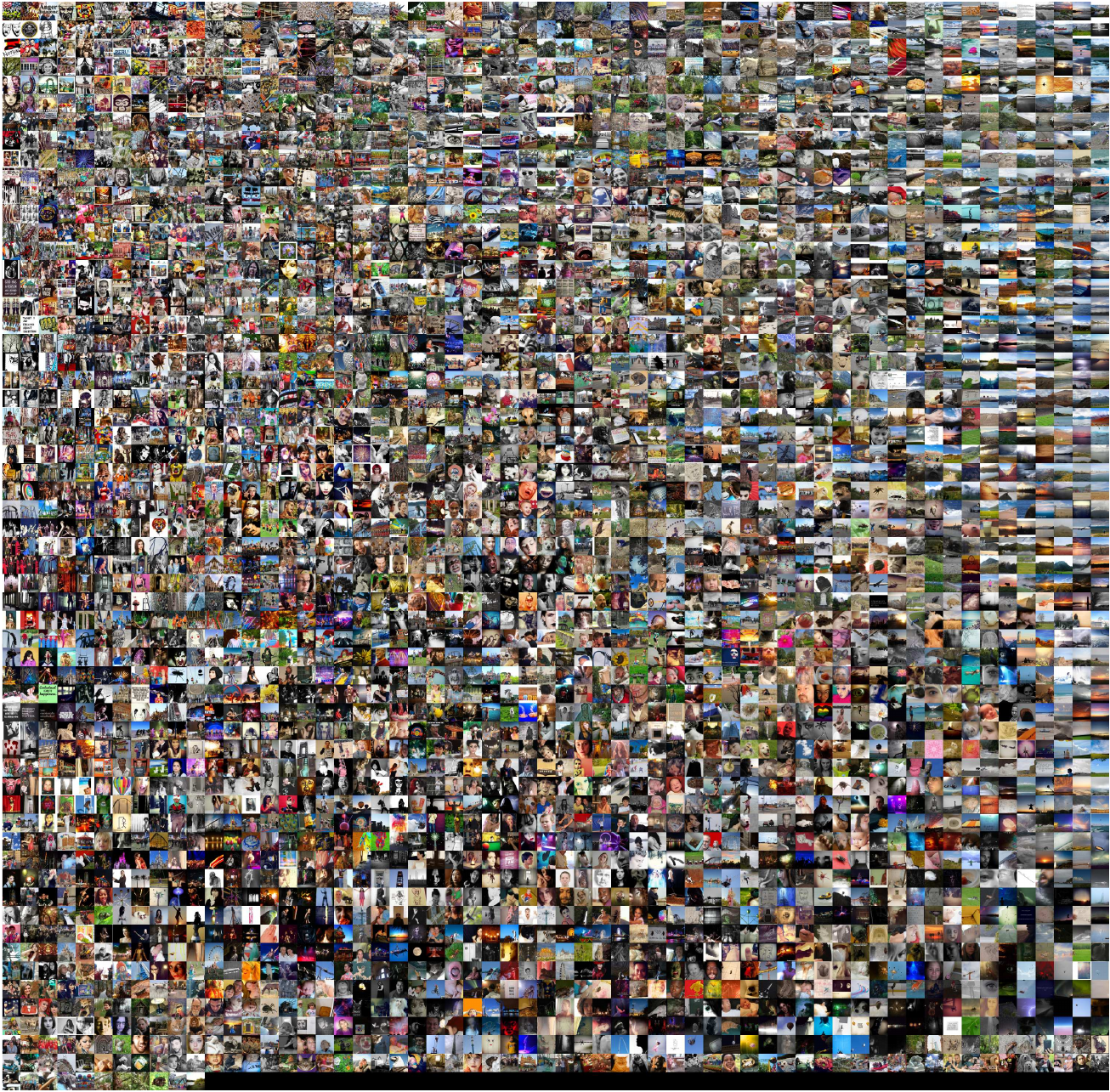}
\label{fig:embedding:bvlc}
}
%\subfigure[Filters of CONV1 Layer from Fine-tuned-CNN]{
%\includegraphics[width=.45\textwidth]{./bvlc_test_vis-crop}
%}
\subfigure[Embedding of the \textit{testing} images using Fine-tuned-CNN features.]{
\includegraphics[width=.4\textwidth]{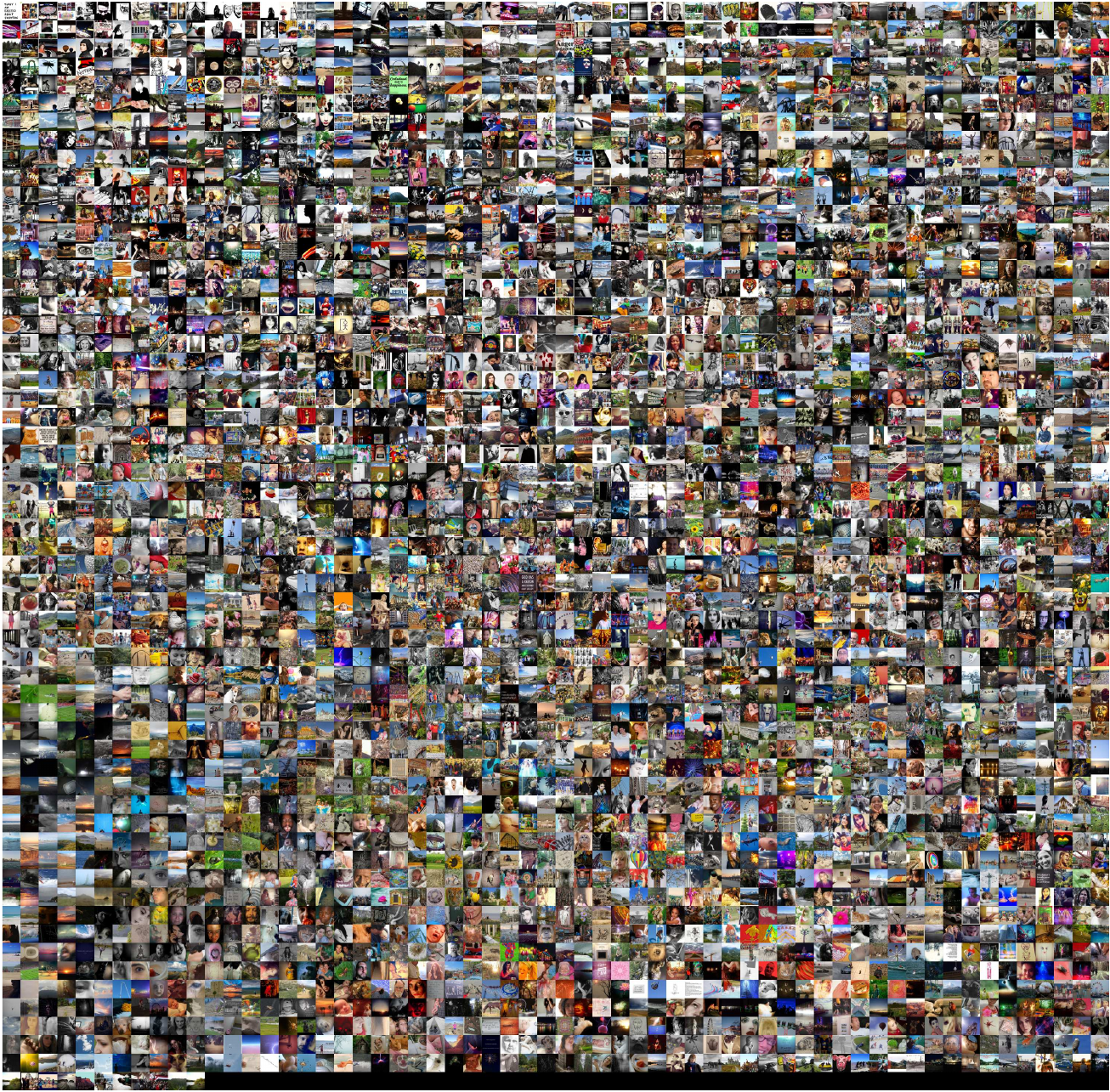}
\label{fig:embedding:fine-tune}
}
%\subfigure[Filters of CONV1 Layer from Fine-tuned-CNN]{
%\includegraphics[width=.45\textwidth]{./amt_test_vis-crop}
%}
\end{center}
\caption{Visualization of learned filters for both ImageNet-CNN and the fine-tuned CNN (best viewed on screen with zoom).}
\label{fig:embedding}
\end{figure*}

\begin{table}[!htpb]
\begin{center}
\begin{tabular}{*{3}{|c}|}
\hline
Algorithms & Correct Samples & Accuracy \\
\hline\hline
ImageNet-CNN & 1120/3490 & 32.1\% \\
Noisy-Fine-tuned-CNN & 1600/3490 & 45.8\% \\
Fine-tuned-CNN & 2034/3490 & 58.3\% \\
\hline
\end{tabular}
\end{center}
\caption{Classification accuracy on the $15\%$ randomly selected testing set labeled by the Amazon Mechanical Turk.}
\label{tab:testing:acc}
\end{table}

We also calculate the confusion matrix of the three algorithms from their prediction results on the testing data to further analyze their performance. \figurename~\ref{fig:confusion:finetune} shows the confusion matrix of the Fine-tuned CNN model. Compared with the other two models, the true negative rates from \textit{Fine-tuned-CNN} are the best in most emotion categories. %In particular, \textit{amusement} and \textit{contentment} are the top two emotions with the largest true positive rates.
Meanwhile, the confusion matrix of \textit{Noisy-Fine-tuned CNN} seems to be more balanced, except for the \textit{contentment} emotion (see\figurename~\ref{fig:confusion:noisyfinetune}). Indeed, these findings are consistent with the number of available labeled samples (see \tablename~\ref{tab:dataset:our}). The more the labeled images, the higher probability that the corresponding emotion will receive a higher true positive rate. \figurename~\ref{fig:confusion:imagenet} shows the confusion matrix using the more general ImageNet-CNN features. It is interesting to see that overall the performance is worse than the Fine-tuned CNN features. However, the true positive rate of \textit{fear} is higher than that of using the Fine-tuned features. %This result suggest that using the general deep visual features is also possible to solve the visual emotion problem.

The embedding of the testing images using deep visual features (we do not show the embedding for Noisy-Fine-tuned-CNN due to space arrangement) is shown in \figurename~\ref{fig:embedding}. The features are also processed using t-SNE~\cite{van2008visualizing}. The embedding using ImageNet-CNN shows that images from the same scene or of similar objects are embedded into neighboring areas. However, the embedding using \figurename~\ref{fig:embedding:fine-tune} seems to make the images more diverse in terms of objects or scenes. This is indeed comply with the fact that even the same object could lead to different visual emotion at its different state, e.g., angry dog and cute dog.
\subsection{Performance of Convolutional Neural Networks on Public Existing Visual Emotion Data Set}

We have described several existing data sets in Section~\ref{sec:dataset}. \tablename~\ref{tab:dataset:base} summarizes the statistics of the three data sets. To the best of our knowledge, no related studies have been conducted on evaluating the performance of Convolutional Neural Networks on visual emotion analysis. In this section, we evaluate all the three deep neural network models on all the three data sets and compare the results with several other state-of-the-art methods on these data sets.

In particular, we extract deep visual features for all the images in the three data sets using the trained deep neural network models from the second to the last layer. In this way, we obtain a $4096$ dimensional feature representation for each image from each deep model. Next, we follow the same evaluation routine described in~\cite{machajdik2010affective} and~\cite{zhou2014learning}. At first, PCA is employed to reduce the dimensions of the features respectively. For all the three data sets, we reduce the number of feature dimensions from $4096$ to $20$, which is capable of keeping at least $90\%$ variance. Next, a linear SVM is trained on the reduced feature space. Following the same experimental approach, the \textit{one v.s. all} strategy is employed to train the classifier. In particular, we randomly split the data into $5$ batches such that $5$-fold Cross Validation is used to obtain the results. Also, we assign larger penalties to \textit{true negative} samples in the SVM training stage in order to optimize the \textit{per class true positive rate} as suggested by both~\cite{machajdik2010affective} and~\cite{zhou2014learning}.

\begin{figure}[!t]
\begin{center}
\includegraphics[width=.45\textwidth]{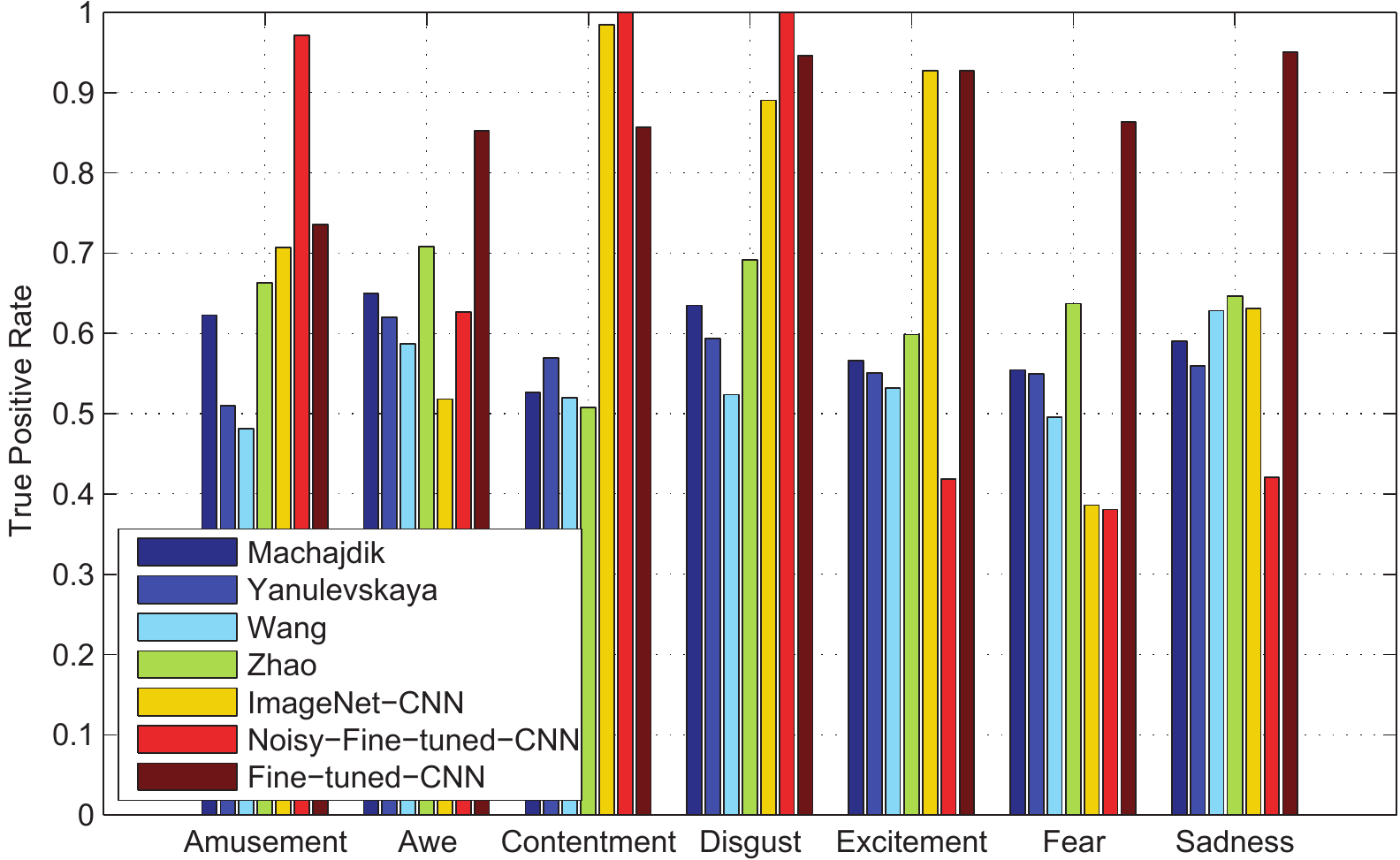}
\end{center}
\caption{Per-class true positive rates of \textit{Machajdik}~\cite{machajdik2010affective}, \textit{Yanulevskaya}~\cite{yanulevskaya2008emotional}, \textit{Wang}~\cite{wei2006image}, \textit{Zhao}~\cite{zhou2014learning}, \textit{ImageNet-CNN}, \textit{Noisy-Fine-tuned-CNN} and \textit{Fine-tuned-CNN} on IAPS-Subset data set.}
\label{fig:iaps}
\end{figure}

\begin{figure}[!t]
\begin{center}
\includegraphics[width=.45\textwidth]{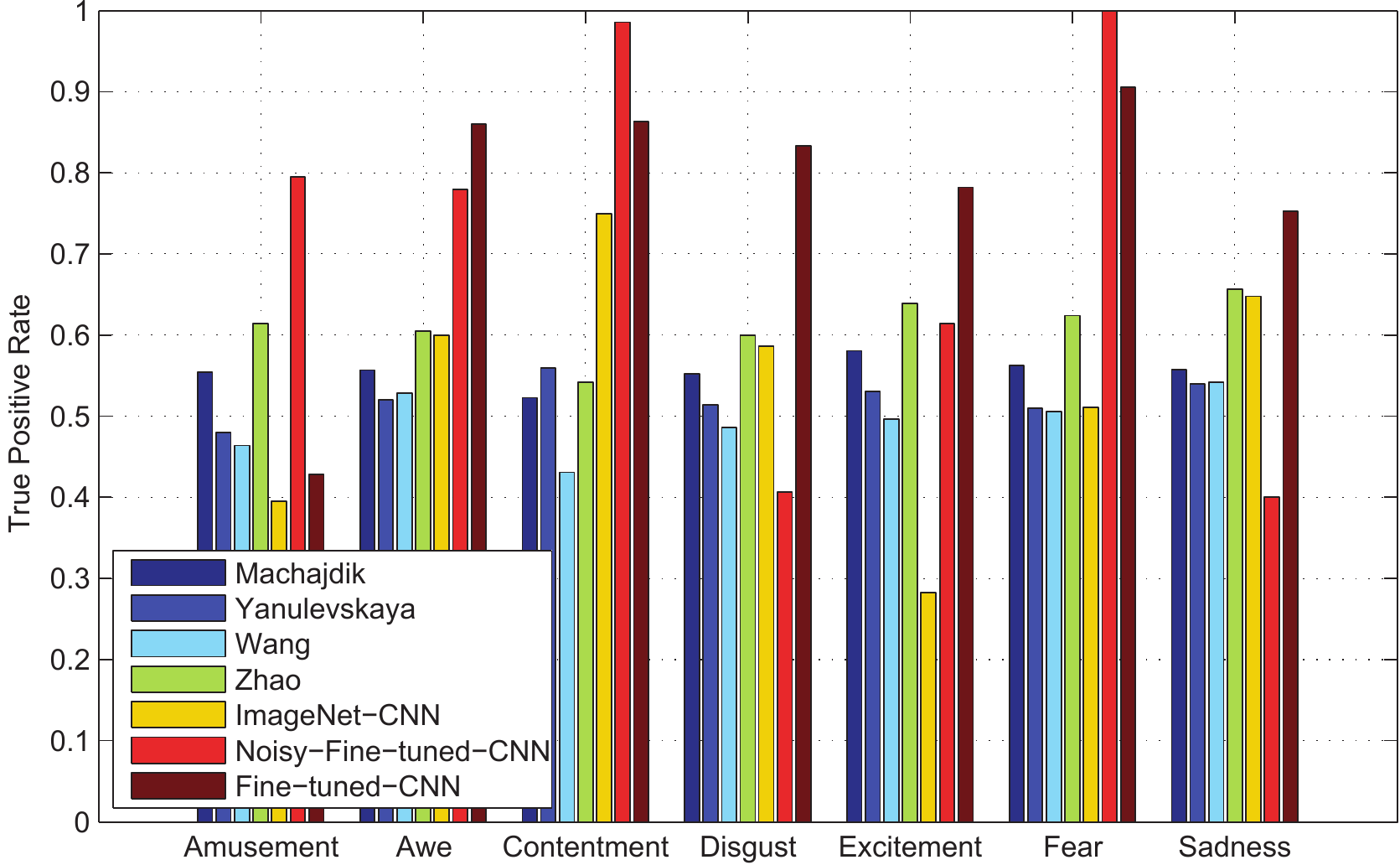}
\end{center}
\caption{Per-class true positive rates of \textit{Machajdik}~\cite{machajdik2010affective}, \textit{Yanulevskaya}~\cite{yanulevskaya2008emotional}, \textit{Wang}~\cite{wei2006image}, \textit{Zhao}~\cite{zhou2014learning}, \textit{ImageNet-CNN}, \textit{Noisy-Fine-tuned-CNN} and \textit{Fine-tuned-CNN} on Abstract Paintings data set.}
\label{fig:abstract}
\end{figure}

\begin{figure}[!t]
\begin{center}
\includegraphics[width=.45\textwidth]{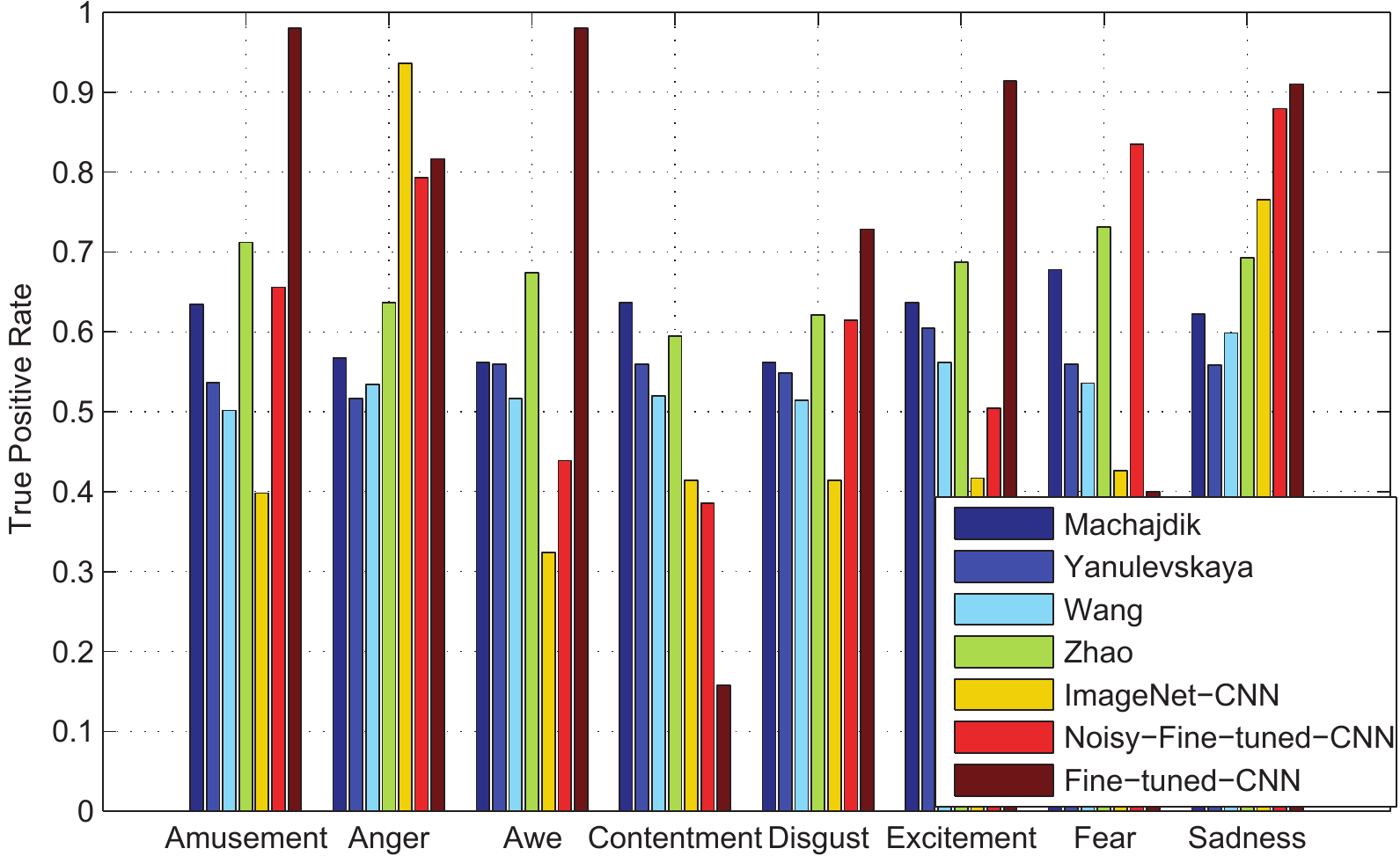}
\end{center}
\caption{Per-class true positive rates of \textit{Machajdik}~\cite{machajdik2010affective}, \textit{Yanulevskaya}~\cite{yanulevskaya2008emotional}, \textit{Wang}~\cite{wei2006image}, \textit{Zhao}~\cite{zhou2014learning}, \textit{ImageNet-CNN}, \textit{Noisy-Fine-tuned-CNN} and \textit{Fine-tuned-CNN} on ArtPhoto data set.}
\label{fig:artphoto}
\end{figure}

We compare the performance of deep features on visual emotion analysis with several other baseline features, including Wang et al.~\cite{wei2006image}, Yanulevskaya~et al.~\cite{yanulevskaya2008emotional}, Machajdik and Hanbury~\cite{machajdik2010affective} and Zhao~et al.~\cite{zhou2014learning}. {\figurename}s ~\ref{fig:iaps},~\ref{fig:abstract} and \ref{fig:artphoto} show the performance of these features on the three data sets respectively. Note that since emotion \textit{anger} only contains $8$ and $3$ images in IAPS-Subset and Abstract Paintings data sets, which are not enough to perform the $5$-fold Cross Validation. We do not report the true positive rates for emotion \textit{anger} on these two data sets.

It is interesting to find out that deep visual features significantly outperform the state-of-the-art manually crafted visual features in some emotion categories. However, the performance of using deep visual features are not consistent across the emotion categories at present. In particular, the performance of directly employing deep visual features from ImageNet-CNN and Noisy-Fine-tuned-CNN differ significantly among categories as well as across data sets. The performance of deep visual features from Fine-tuned-CNN is relatively more consistent. However, it has poor performance on emotions \textit{Contentment} and \textit{Fear} in the ArtPhoto data. These results suggest that it is still challenging to solve visual emotion analysis even with the state-of-the-art deep visual features. Meanwhile, the performance of deep visual features also suggests the promise of using CNNs in visual emotion analysis. Overall, this may encourage the development of more advanced deep architectures for visual emotion analysis, as well as development of other approaches.

%In almost all the testing cases, the features inspired from color and art can only achieve a per-class true positive rates of less than $70\%$. However, for both deep visual features from the Convolutional Neural Networks, they can achieve over $90\%$ per-class true positive rates. In particular, in the ArtPhoto data set, since all the emotion categories have a relatively large number of images, both different groups of deep features outperform the low-level features in all emotion categories. Meanwhile, the fine-tuned CNN features perform better than all the low-level features on all the three data set, where CNN may become worse on some emotion categories. This may suggest that the fine-tuned deep visual features are more appropriate for visual emotion analysis than the more general object-oriented ImageNet-CNN features.

\section{Conclusions}
\label{sec:conclusion}
In this work, we introduce the challenging problem of visual emotion analysis. Due to the unavailability of a large scale well labeled data set, little research work has been published on studying the impact of Convolutional Neural Networks on visual emotion analysis. In this work, we are introducing such a data set and intend to release the data set to the research community to promote the research on visual emotion analysis with the deep learning and other learning frameworks. Meanwhile, we also evaluate the deep visual features extracted from differently trained neural network models. Our experimental results suggest that deep convolutional neural network features outperform the state-of-the-art hand-tuned features for visual emotion analysis. In addition, fine-tuned neural network on emotion related data sets can further improve the performance of deep neural network. Nevertheless, the results obtained in this work are only a start for the research on employing deep learning or other learning frameworks for visual emotion analysis. We will continue the collection of labeled data from AMT with a plan to submit additional 1 million images for labeling. We hope our visual emotion analysis results can encourage further research on online user generated multimedia content in the wild. Better understanding the relationship between emotion arousals and visual stimuli and further extending the understanding to valence are the primary future directions for visual emotion analysis.

%Understanding the relationship between emotion arousals and visual stimuli and extending the understanding to valence are the primary future directions for visual emotion analysis.

\section*{Acknowledgement}
This work was generously supported in part by Adobe Research, and New York State CoE IDS. We thank the authors of~\cite{Zhao:2014:EPF:2647868.2654930} for providing their algorithms for comparison.

\bibliographystyle{aaai}
\bibliography{aaai_2016_742}
\end{document}